# A Study of Social and Behavioral Determinants of Health in Lung Cancer Patients Using Transformers-based Natural Language Processing Models

Zehao Yu[1], Xi Yang[1,3], Chong Dang[1], Songzi Wu[1], Prakash Adekkanattu[4], Jyotishman Pathak[5], Thomas J. George[2], William R. Hogan[1], Yi Guo[1,3], Jiang Bian[1,3], Yonghui Wu[1,3]

[1]Department of Health Outcomes and Biomedical Informatics, [2]Division of Hematology & Oncology, Department of Medicine, College of Medicine, [3]Cancer Informatics Shared Resources, University of Florida Health Cancer Center, University of Florida, Gainesville, Florida, USA; [4]Information Technologies and Services, [5]Department of Population Health Sciences, Weill Cornell Medicine, New York, NY, USA.

**Abstract**

*Social and behavioral determinants of health (SBDoH) have important roles in shaping people's health. In clinical research studies, especially comparative effectiveness studies, failure to adjust for SBDoH factors will potentially cause confounding issues and misclassification errors in either statistical analyses and machine learning-based models. However, there are limited studies to examine SBDoH factors in clinical outcomes due to the lack of structured SBDoH information in current electronic health record (EHR) systems, while much of the SBDoH information is documented in clinical narratives. Natural language processing (NLP) is thus the key technology to extract such information from unstructured clinical text. However, there is not a mature clinical NLP system focusing on SBDoH. In this study, we examined two state-of-the-art transformer-based NLP models, including BERT and RoBERTa, to extract SBDoH concepts from clinical narratives, applied the best performing model to extract SBDoH concepts on a lung cancer screening patient cohort, and examined the difference of SBDoH information between NLP extracted results and structured EHRs (SBDoH information captured in standard vocabularies such as the International Classification of Diseases codes). The experimental results show that the BERT-based NLP model achieved the best strict/lenient F1-score of 0.8791 and 0.8999, respectively. The comparison between NLP extracted SBDoH information and structured EHRs in the lung cancer patient cohort of 864 patients with 161,933 various types of clinical notes showed that much more detailed information about smoking, education, and employment were only captured in clinical narratives and that it is necessary to use both clinical narratives and structured EHRs to construct a more complete picture of patients' SBDoH factors.*

**Introduction**

People's health outcomes are associated with complex, multi-level factors including clinical, biological, social, and health behaviors. To measure health outcomes, transdisciplinary approaches that can leverage a wide range of potential factors are needed. Major initiatives such as the Patient-Centered Outcomes Research Institute (PCORI)-funded PCORnet[1] and the Observational Health Data Sciences and Informatics[2] (OHDSI) have been established to promote research using electronic health records (EHRs) for health outcomes research. Clinical factors (e.g., diagnoses, medications) captured in EHRs have been widely used in various clinical studies. Due to lack of resources, patient's biological, social and health behaviors are under-studied in health outcomes-related studies, which may cause potentially confounding issues (in statistical analyses) or misclassification errors (in machine-learning based classifiers). The Electronic Medical Records and Genomics[3] (eMERGE) network and the NIH All of Us Research program have been established to develop resources of genetic data linked with EHRs. On the other side, through well-established conceptual frameworks such as the social-ecological model[4] and the NIMHD Minority Health and Health Disparities Research Framework[5], patient's social determinants of health (SDoH; e.g., education, employment, income disparities) and their health behaviors (or behavioral determinants of health – BDoH; e.g., smoking, alcohol use) are increasingly recognized as important factors influencing health outcomes.[6] There is an increasing interest to examine social and behavioral determinants of health (SBDoH) in shaping people's health.

SBDoH are important risk factors affecting people's health and healthcare outcomes. The most consistent predictors for the likelihood of death in any given year is level of education[7] and poverty, estimated to account for 6% of US mortality[8]. In cancer, the second leading cause of death in US, up to 75% of cancers occurrences are associated with SBDoH factors[9]. Studies have reported that many SBDoH contribute to individual cancer risk, influence the likelihood of survival, and affect cancer early prevention and health equity.[10–12] A recent study[13] reported that SDoH factors such as poverty, lack of education, neighborhood disadvantage, and social isolation play important roles in breast cancer stage and survival. Many SBDoH factors are also associated with the screening of cervical cancer, breast cancer, and lung cancer.[14] Reports from US institutes (e.g., Institute of Medicine[15], HealthyPeople 2020[16], and Health and Human Services[17]) and international organizations (e.g., World Health Organization) reflect the increasing consensus of acknowledging SBDoH as significant contributory factors. The World Health Organization (WHO) defined structured codes to capture some of the SBDoH in EHRs. One potential source of SBDoH in the EHRs is in the International Classification of Diseases, Tenth Revision, Clinical Modification (ICD-10-CM) Z codes (Z55–Z65). In February 2018, ICD-10-CM Official Guidelines for Coding and Reporting approved that all clinicians, not just the physicians, involved in the care of a patient can document SBDoH using these Z codes. In our previous study[18], we have examined the use of ICD10 Z codes in the OneFlorida Clinical Research Consortium and reported a low rate of utilization for these Z codes (270.61 per 100,000 at the encounter level and 2.03% at the patient level). In order to advance cancer control and prevention, a transdisciplinary approach that integrates biological, clinical, social, and behavioral factors is needed. Nonetheless, SBDoH factors are scarcely and inconsistently documented in structured EHRs despite EHR systems provided opportunities for them to be manually entered as discrete data, but are often available in clinical text. Natural language processing (NLP) systems that systematically extract SBDoH factors from unstructured clinical text are needed to better assess cancer outcomes.

NLP is the key technology to unlock this critical information embedded in clinical narratives to support various downstream clinical studies that depend on structured data.[19,20] NLP has received great attention in recent years. Various clinical NLP systems, such as MedLEE[21] (Medical Language Extraction and Encoding System), MetaMap[22], KnowledgeMap[23] and cTAKES[24] (clinical Text Analysis and Knowledge Extraction System), have been developed to extract medical concepts from clinical narratives. To ensure the accuracy of information extraction, researchers have invested great effort into the development of clinical NLP methods.[25,26] Clinical NLP systems approach the extraction of medical concepts from clinical narratives as a Named-Entity Recognition[19] (NER) task. NER first identifies the boundaries (start position and end position in text) of medical concepts and then determines their semantic categories (e.g., diseases, medications). NLP methods based on statistical machine learning (ML) models have been increasingly applied and demonstrated good performance. Recently, NLP methods based on deep learning (DL) models have demonstrated superior performance than traditional ML models.[27–30] However, most of the NLP systems focused on clinical factors (medical concepts directly generated by clinical practice, e.g., diseases and medications); NLP methods to extract SBDoH [16,31–33] factors have been under-studied.

There are limited studies exploring NLP methods to extract SBDoH from clinical narratives. Feller et al[34] developed machine-learning classifiers to determine whether 11 categories of SBDoH are presented in clinical documents. Stemerman et al[35] also compared 5 machine-learning classifiers in classifying whether or not 5 categories of SBDoH presented in clinical text. Recently, Lybarger et al[36] developed an SBDoH corpus that consisted of 4,480 social history sections and applied deep learning-based NLP methods to extract 12 categories of SBDoH. To the best of our knowledge, no study has been reported investigating the difference between SBDoH extracted from clinical narratives and those captured in structured EHRs. This study aims to explore a state-of-the-art NLP model, clinical transformers, to extract SBDoH from clinical narratives. Using cancer as a study case, we (1) systematically examined two state-of-the-art transformer-based NLP models in extraction SBDoH from clinical narratives, (2) applied a transformer-based NLP model to extract SBDoH concepts for a lung cancer screening patient cohort, and (3) examined the difference between SBDoH extracted by NLP and those captured in structured EHRs. This is one of the earliest studies to apply transformer models to extract SBDoH and examine the difference between clinical narratives and structured EHRs.

**Methods**

**Dataset**

This study used EHR data from the University of Florida (UF) Health Integrated Data Repository (IDR) from 1999 to 2020, including both structured data and clinical narratives. Supported by the UF Clinical and Translational Institute (CTSI) and the UF Health, the UF Health IDR is a clinical data warehouse (CDW) that aggregates data from the university's various clinical and administrative information systems, including the Epic (Epic Systems Corporation) electronic medical record (EMR) system. This study was approved by the UF Institutional Review Board (IRB201902362).

**Study design**

**Identify cancer patient cohorts**: Using UF IDR, we identified a general cancer (GC) cohort and a lung cancer screening (LCS) cohort. The GC cohort consists of 20,000 cancer patients sampled using a stratified random sampling (by cancer types), which has a total number of 1.5 million clinical notes. The LCS cohort was identified using the following rule-based phenotyping method:

(1) The age at the first low-dose computed tomography (LDCT) date is between 50 and 80;

(2) Most recent smoking status is Never smoker, but there's evidence on patient ever smoked within the timeframe OR Most recent smoking status is Current smoker OR most recent smoking status is Former smoker, and at least one most recent quit year in structured data or note <= 15 years OR most recent smoking status is Former smoker but has Current smoker record in structured data within the timeframe; and

(3) Majority smoking pack-year record in structure and note with desired encounter type>= 20.

The LCS cohort consists of a total of 864 lung cancer patients with a total number of 161,933 various types of clinical notes. There are no patients existed in both cohorts.

**Training and test datasets**: We used the GC cohort to train various NLP methods, optimize model parameters. To develop a training corpus for SBDoH, we developed a filtering pipeline using note types (identified by domain experts) and a total number of 30 keywords to identify clinical notes potentially with rich content of SBDoH concepts. Using the filtering pipeline, we identified a total of 225,441 notes potentially with rich SBDoH concepts from the GC cohort, where 500 notes were randomly sampled for annotation. Two annotators manually identified all SBDoH concepts according to predefined guidelines. After annotation, we divided the 500 notes into a training set of 400 notes – used to develop various NLP models, and a test set of 100 notes – used to evaluate the performance of NLP models and select the best model. We trained various transformer-based NLP models using the training set and evaluated the performance using the test set (in terms of F1-score). Then, we identified the best NLP method according to the performance on test set and applied this model to extract SBDoH concepts from all clinical notes of the LCS cohort. To compare the NLP extracted SBDoH with those captured in structured EHRs, we developed a normalization pipeline to align the NLP results into predefined categories from the structured EHRs. Figure 1 shows an overview of the study design.

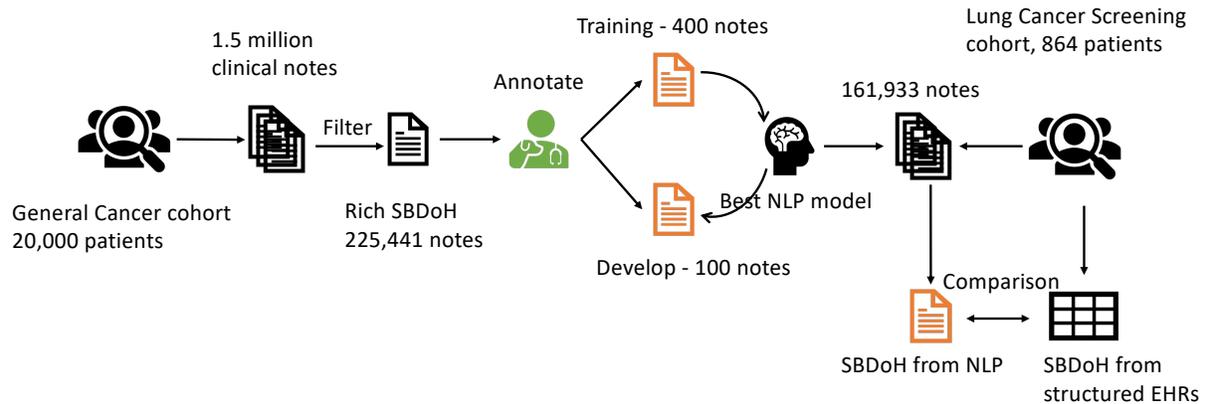

**Figure 1.** An overview of study design.

### NLP methods

We explored two state-of-the-art transformer-based NLP methods, including (BERT)[37] and (RoBERTa)[38], and compared them with a widely used deep learning model based on Long short-term memory (LSTM)[39]. Previously, we have developed clinical transformer package[40] based on the transformer architectures implemented in the HuggingFace[41] in PyTorch[42]. We used the BERT model and RoBERTa model implemented in our clinical transformer package to develop transformer-based NLP solutions for SBDoH as they achieved the top 2 best performances in our previous study[40]. For BERT, we used the 'base' setting in this study. Following our previous studies[40,43,44] on clinical transformers, we also examined pre-trained transformers from general English corpus (denoted as '_general', e.g., 'BERT_general') and clinical transformers pre-trained using clinical notes from the Medical Information Mart for Intensive Care III (MIMIC-III) database[45] (denoted as '_mimic', e.g., 'BERT_mimic'). We adopted the default parameters optimized in our clinical transformer package[40].

### SBDoH from structured EHRs

We collected 5 categories of structured SBDoH from the structured tables in the UF IDR database. At UF health IDR database, gender and ethnicity were captured as patient-level information, whereas the smoking status was captured for each encounter (one patient may have multiple smoking status from different encounters). For education and employment, there is no information captured in structured tables.

**NLP results alignment**

Our NLP system identified SBDoH concepts at the document-level, which is different from the structured SBDoH information that was captured at patient-level or encounter-level. In order to compare NLP results with structured EHRs, we developed a rule-based pipeline to align the document-level NLP results to structured EHRs. More specifically, we adopted a majority voting strategy to aggregate multiple mentions of gender and ethnicity from clinical notes into patient-level categories - the most frequent category will be used. Smoking status is a factor that may change over time, therefore, we compared smoking status at a "per patient / per year" setting. Under the "per patient / per year" setting, a patient is supposed to have a smoking status datapoint for each year in the EHR database – e.g., a patient with 5-year history is supposed to have 5 smoking datapoints, one for each year. We aggregated smoking information from both the structured EHRs (encounter-level) and NLP (document-level) into the "per patient/per year" setting for comparison. Similarly, the majority voting strategy was used. According to the structured smoking categories defined in structured EHRs, there are 7 different smoking categories. To algin the NLP extracted smoking information to the 7 structured categories, we first generated a list of unique smoking status captured by NLP and then manually reviewed them (a total of 429) to map them into one of the 7 structured categories. As employment status and education were not captured in structured EHR, we were not able to compare.

**Evaluation**

We used both strict (i.e., the beginning and end boundaries of a concept have to be exactly the same with gold-standard annotation) and lenient precision, recall, and F1-score to evaluate our NLP systems. Precision is defined as *(the number of SBDoH concepts correctly identified by the NLP system) / (total number of concepts identified by NLP)*; recall is defined as *(the number of SBDoH concepts correctly identified by the NLP system) / (total number of concepts annotated by experts)*; F1-score is defined as *"(2\*precision\*recall)/(precision+recall)"*. To compare NLP results with structured data, we report the number of patients with structured SBDoH *vs* the number of patients with NLP extracted SBDoH; the number of patients who only have structured information *vs* the number of patients who only have NLP extracted information.

**Results**

Two annotators annotated 1,876 SBDoH concepts from 500 clinical notes. The inter-annotator agreement measured by token level kappa score with 40 overlapped clinical notes was 0.97, indicating the two annotators have a good agreement in the annotation. Table 1 shows the distribution of SBDoH concepts in the training and test set.

**Table 1.** Number of SBDoH concepts in the training set and test set.

| Entity type | Training set | Test set |
|---|---|---|
| Gender | 441 | 120 |
| Ethnicity | 21 | 6 |
| Smoking | 665 | 178 |
| Employment | 27 | 10 |
| Education | 335 | 73 |

**Table 2.** Comparison of performance for BERT, RoBERTa, and LSTM-CRFs on the test set.

| | Strict | | | Lenient | | |
|---|---|---|---|---|---|---|
| Model | Precision. | Recall | F1 | Precision | Recall | F1 |
| BERT_general | 0.8848 | 0.8734 | **0.8791** | 0.9058 | 0.8941 | **0.8999** |
| BERT_mimic | 0.8952 | 0.8605 | 0.8775 | 0.9086 | 0.8734 | 0.8906 |
| RoBERTa _general | **0.9017** | 0.8295 | 0.8641 | **0.9242** | 0.8501 | 0.8856 |
| RoBERTa _mimic | 0.8914 | 0.8269 | 0.8579 | 0.9220 | 0.8553 | 0.8874 |
| LSTM_general | 0.7519 | 0.9327 | 0.8326 | 0.7700 | 0.9551 | 0.8526 |
| LSTM_mimic | 0.7674 | **0.9369** | 0.8438 | 0.7804 | **0.9527** | 0.858 |

**Table 3.** Detailed performance for each SBDoH category for the best NLP model - BERT_general.

| BERT_general | Strict | | | Lenient | | |
|---|---|---|---|---|---|---|
| Type | Precision | Recall | F1-score | Precision | Recall | F1-score |
| Gender | 0.9091 | 0.9167 | 0.9129 | 0.9091 | 0.9167 | 0.9129 |
| Ethnicity | 0.8571 | 1.0000 | 0.9231 | 0.8571 | 1.0000 | 0.9231 |
| Education | 0.8857 | 0.8493 | 0.8671 | 0.9286 | 0.8904 | 0.9091 |
| Smoking | 0.8764 | 0.8764 | 0.8764 | 0.9045 | 0.9045 | 0.9045 |
| Employment | 0.6667 | 0.4000 | 0.5000 | 0.6667 | 0.4000 | 0.5000 |
| Overall | 0.8848 | 0.8734 | 0.8791 | 0.9058 | 0.8941 | 0.8999 |

Table 2 compares two transformer-based NLP methods with a widely used deep learning model – LSTM-CRFs in extracting SBDoH concepts from clinical narratives. Both two transformer-based NLP methods outperformed LSTM-CRFs. Among the two transformer-based NLP methods, the BERT_general model achieved the best strict/lenient F1-score of 0.8791 and 0.8999 on the test set, respectively. Table 3 shows the detailed performance for each of the 5 SBDoH categories for the best NLP model. We applied the best NLP model, BERT_general, to extract SBDoH concepts from all clinical notes of the LCS patient cohort and compared them with structured EHRs. Table 4 shows the comparison results between NLP and structured EHRs. For the 864 patients in LCS cohort, the structured EHRs covered 99.65% patients for gender and ethnicity. However, the structured EHRs only covered 56.55% of smoking datapoints *vs* 71.57% from NLP. There is no information captured in structured EHRs for education and employment. Information about education and employment was documented in the narrative clinical text (39.35% for education and 47.22% for employment).

**Table 4. Comparison of NLP extracted concepts with structured concepts for LCS cancer patients**

| | Gender | Ethnicity | Smoking* | Education | Employment |
|---|---|---|---|---|---|
| #concepts detected by NLP | 88,015 | 14,866 | 104,201 | 22,460 | 2,236 |
| #patients with NLP detected concepts | 861(99.65%) | 713(82.52%) | 5,736(71.57%) | 340(39.35%) | 408(47.22%) |
| #patients with structured concepts | 861(99.65%) | 861(99.65%) | 4,524(56.44%) | 0 | 0 |
| #patients with NLP consistent with structured concepts | 860(99.53%) | 703(81.37%) | 3,015(37.62%) | 0 | 0 |
| #patients only have NLP concepts | 2(0.23%) | 0 | 1,517(18.92%) | 340(39.35%) | 408(47.22%) |
| #patients only have structured concepts | 2(0.23%) | 150(17.36%) | 308(3.84%) | 0 | 0 |

*For smoking, the numbers and proportions were calculated using datapoints – the 864 patients are supposed to have 104,201 datapoints using "per patient / per year" setting. The proportions for other SBDoH concepts were calculated according to a total number of 864 patients in LCS cohort.

## Discussion and conclusion

SBDoH are important factors associated with cancer risk, prevention, screening, and survival. This study applied state-of-the-art transformer-based NLP methods to extract SBDoH concepts from clinical narratives and took lung cancer as a study case to compare the difference between clinical narratives and structured EHRs. Our experimental results show that two transformer-based NLP methods, BERT and RoBERTa, outperformed a widely used deep learning model – LSTM-CRFs – in extracting SBDoH from clinical narratives. Among the 4 transformer models, the BERT_general model pretrained using general English corpora achieved the best strict/lenient F1-score of 0.8791 and 0.8999, respectively, indicating the efficiency of transformer-based NLP methods. The BERT_general model pretrained using general English corpora outperformed another BERT_mimic model pretrained using MIMIC III clinical text, which is consistent with our previous study of using transformer models for de-identification of clinical notes.[46] Similar to the protected health information in the de-identification task, the SBDoH concepts are closer to the general English language than the medical language. Therefore, the transformer models pretrained with a large volume of general English text are better in recognizing those SBDoH concepts closer to general English.

We examined 5 categories of SBDoH concepts in a lung cancer patient cohort. The best NLP model, BERT_general, was used to extract SBDoH concepts from all clinical notes of the LCS patient cohort. The comparison between NLP extracted results and structured EHRs shows that the structured EHRs have good coverage (>99% patients) for gender and ethnicity; but for smoking, education, and employment, much detailed information was documented in clinical narratives. For smoking, 71.57% of datapoints were documented in clinical notes, whereas, only 56.44% were captured in structured EHRs. Between the smoking datapoints extracted by NLP and structured EHRs, there are only 37.62% consistent with each other; there are 18.92% of smoking datapoints only from NLP, indicating the necessity of using both NLP and structured EHRs for cancer-related studies. For education and employment, currently there was no information captured in structured EHRs; but we can get 39.35% education information and 47.22% employment information from clinical notes using NLP. Smoking, education, and employment are important factors associated with many cancers. NLP systems could fill the gap of using these SBDoH factors in clinical studies. Smoking is an important BDoH factor for lung cancer. To better understand how smoking was documented in clinical narratives and structured EHRs over time, we calculated the normalized proportion of patients with smoking information for each year (defined as "*[number of patients with smoking information in one year] / [total number of patients in that year]*") and plotted the curve from 2009 to 2020 in Figure 2 (structured smoking information showed up in the LCS cohort starting from 2009; the EHR data for 2020 is not complete as we queried data in mid 2020). From Figure 2 we can see that smoking information is consistently documented in clinical text, whereas, the proportion of patients with smoking information in structured EHRs was low in the beginning but increased a lot over time. This increase was likely a result of institutional requirements for tobacco use documentation as part of quality assurance programs and lung cancer screening program compliance. Note that Figure 2 only shows the proportion of patients with smoking information or not, it can't be used to assess how complete it is in structured EHRs. In fact, we can see from Table 4 that much of the detailed smoking datapoints were still documented in clinical text than structured EHRs when compared using the "per patient / per year" setting.

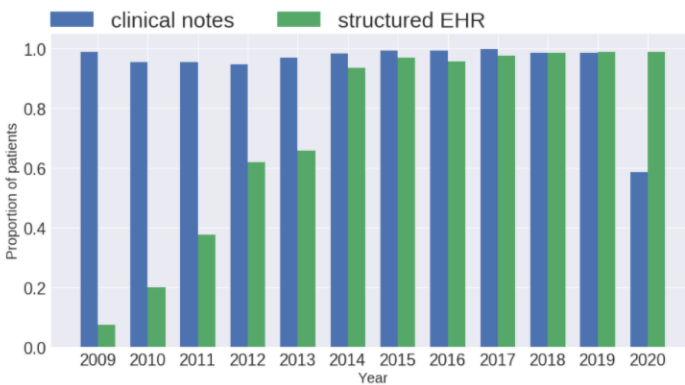

**Figure 2.** A comparison of patients' smoking status between clinic narratives and structured EHRs.

NLP is the key technology to extract SBDoH concepts from clinical narratives. Yet, most of the NLP studies focused on clinical factors such as diseases, medications; the application of NLP to extract SBDoH concepts is under-studied. Clinical NLP systems that can accurately identify SBDoH concepts are needed to fill the gap of using these factors in various health outcomes studies. The ultimate goal of this project is to develop a clinical NLP package that can accurately identify a wide range of SBDoH concepts from clinical narratives and populate them into structured EHR databases such as the NLP tables defined in the OHDSI common data model (CDM) and PCORnet CDM.

**Limitations and future plans**

This study has limitations. As a preliminary study, we limited SBDoH concepts to 5 categories including gender, ethnicity, smoking, education, and employment. Future studies need to examine more SBDoH categories associated with health outcomes. We took lung cancer as a study case to examine how current NLP models could identify SBDoH concepts and how complete they were captured in NLP *vs* structured EHRs. The documentation of SBDoH concepts might be different in other disease domains. During the time of this study analysis, the institutional policy changed to require more discrete data collection related to smoking use, which can impact the results. We expect more studies to examine the use of SBDoH in various diseases. Also, this study examined clinical notes from a single site, future work should examine the proposed transformer-based NLP system at cross-institute settings.

**Acknowledgement**

This study was partially supported by a Patient-Centered Outcomes Research Institute® (PCORI®) Award (ME-2018C3-14754), a grant from National Institute on Aging 1R56AG 069880, a grant from the National Cancer Institute, 1R01CA246418 R01, a grant from CDC (Centers for Disease Control and Prevention) 1U18DP006512-01, the University of Florida (UF) SEED Program (DRPD-ROF2020, P0175580), and the Cancer Informatics and eHealth core jointly supported by the UF Health Cancer Center and the UF Clinical and Translational Science Institute. The content is solely the responsibility of the authors and does not necessarily represent the official views of the funding institutions.